\documentclass[conference]{IEEEtran}
\usepackage{amsmath,amssymb}
\usepackage{booktabs}
\usepackage{graphicx}
\usepackage{xcolor}
\usepackage{url}
\usepackage[hidelinks]{hyperref}
\usepackage{tikz}
\usetikzlibrary{arrows.meta,positioning,fit,shapes.geometric}
\graphicspath{{figures/}}
\newcommand{\calm}{\textsc{CALM}}
\newcommand{\jsd}{D_{\mathrm{JS}}}
\newcommand{\wone}{W_1}

\title{\calm: A Calibrated LLM--Choice--Network Framework for Activity-Based Traveler Simulation}

\author{\IEEEauthorblockN{Yezhou Cheng}
\IEEEauthorblockA{\textit{Independent Researcher}}}

\begin{document}
\maketitle

\begin{abstract}
We present \calm, a reproducible hybrid framework that integrates an optional large language model (LLM) activity planner with calibrated stochastic choice, shared network feedback, memory/habit, typed feasibility checks, and deterministic offline replay. Unlike trip-mode classifiers or diary-only generators, \calm\ executes a closed traveler-day loop and evaluates each generative module against an empirical, reproducible baseline. On the 2024 New York City Citywide Mobility Survey (CMS), 110,691 seven-mode trips are split by respondent into 78,487 training and 32,204 holdout trips. Training-only alternative-specific constant calibration reduces mean holdout mode Jensen--Shannon divergence from 0.15599 to 0.00394 across ten seeds. A matched live-LLM ablation then quantifies trade-offs among aggregate fit, temporal fit, behavioral persistence, and feasibility, while frozen prompt--response pairs support deterministic replay of downstream simulation. Controlled weather, delay, fare, and parking ladders further demonstrate consistent and interpretable responses under intervention. \calm\ contributes a reproducible protocol for integrating and evaluating generative planners in traveler simulation through person-disjoint calibration, matched module ablation, controlled stress testing, and end-to-end traceability.
\end{abstract}


\section{Introduction}
Activity-based travel models treat daily mobility as interdependent activity and travel decisions rather than independent trips \cite{bowman2001activity,bhat2003activity}. Agent-based systems such as ALBATROSS and MATSim operationalize this view with explicit travelers and demand--network interaction \cite{arentze2000albatross,horni2016matsim}. Their policy usefulness depends on calibration: matching aggregates while preserving decision structure remains hard \cite{agriesti2023bayesian,schultz2022bayesian}.

LLM agents expand the design space. Generative architectures can produce fluent plans and persistent narratives \cite{park2023generative,argyle2023out}; mobility systems use them for schedules, trajectories, or traveler decisions \cite{wang2024urban,ju2025trajllm,liu2025toward,liu2024limited}. Persona alignment can improve a defined choice task \cite{liu2026aligning}, and hybrid platforms embed LLM semantics inside validated urban pipelines \cite{santos2026citybehavex}. At the same time, reviews stress that face validity is not empirical validity \cite{larooij2026validation,gao2026mobilitysurvey}, and plausible mobility narratives can still mismatch spatial, temporal, and OD distributions \cite{santos2026plausible}.

\calm\ is a calibrated hybrid framework for activity-based traveler simulation. It supports traveler-day activity planning and enables controlled comparisons of planner, memory, and choice mechanisms under matched travel contexts. Using New York City CMS data \cite{nycdot2024cms}, this paper makes four contributions. First, it introduces a modular closed loop---planner $\to$ feasibility $\to$ choice $\to$ network $\to$ memory---that accommodates fixed, mock, or live-LLM planners behind a common typed interface. Second, it establishes a person-disjoint empirical anchor through training-only survey-weighted ASC calibration and frozen holdout evaluation. Third, it provides a matched evaluation protocol that separates aggregate fit, live-module effects, controlled stress response, and supervised prediction into task-appropriate experiments. Fourth, it adds a reproducibility layer with prompt hashes, revision traces, deterministic repair, and deterministic offline replay from frozen LLM responses. Together these components turn generative traveler planning into a controlled, traceable, and reproducible simulation module rather than a standalone narrative generator.

\section{Related Work and Comparison Frame}
\subsection{Traditional activity-based and agent-based models}
Classical ABM/ABS systems emphasize constraint-consistent activity patterns and network coupling \cite{bowman2001activity,bhat2003activity,arentze2000albatross,horni2016matsim}. Calibration is often treated as a large-scale optimization problem \cite{agriesti2023bayesian,schultz2022bayesian}. \calm\ combines the choice--network loop with survey-weighted ASC calibration, providing a transparent empirical reference for evaluating generative components.

\subsection{LLM mobility systems}
Recent work explores LLM residents and trajectory agents \cite{wang2024urban,ju2025trajllm}, limited-information mobility modeling \cite{liu2024limited}, conceptual LLM--ABM transportation frameworks \cite{liu2025toward,liu2025llmabm}, and persona-aligned travel choice \cite{liu2026aligning}. Evaluation studies and surveys identify calibration, reproducibility, and task-aligned validation as central requirements \cite{larooij2026validation,santos2026plausible,gao2026mobilitysurvey}. These strands motivate a framework that connects traveler-day generation with empirical calibration, controlled intervention, and reproducible replay in a single experimental workflow.

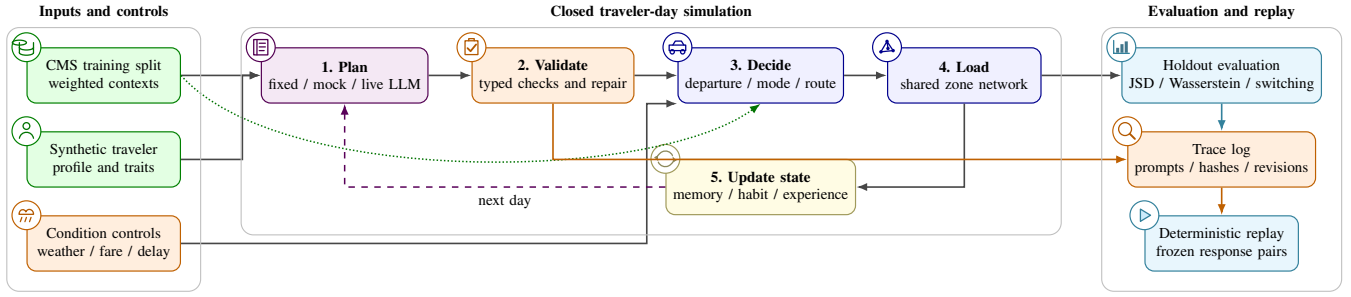
\begin{figure*}[!t]
\centering
\resizebox{0.98\textwidth}{!}{%
\begin{tikzpicture}[
  node distance=4.5mm and 7mm,
  font=\scriptsize,
  module/.style={draw=blue!55!black,rounded corners=1.5mm,align=center,
    minimum height=9mm,minimum width=25mm,fill=blue!6,line width=.55pt},
  planner/.style={module,fill=violet!9,draw=violet!65!black},
  gate/.style={module,fill=orange!13,draw=orange!75!black},
  data/.style={module,fill=green!11,draw=green!50!black},
  output/.style={module,fill=cyan!8,draw=cyan!55!black},
  memory/.style={module,fill=yellow!13,draw=yellow!55!black},
  iconbg/.style={draw=#1,fill=white,line width=.55pt},
  iconline/.style={draw=#1,line width=.6pt,line cap=round,line join=round},
  group/.style={draw=black!25,rounded corners=2mm,
    inner sep=3.2mm,line width=.5pt},
  arr/.style={-{Latex[length=1.8mm]},line width=.75pt,draw=black!72},
  feedback/.style={-{Latex[length=1.8mm]},line width=.7pt,
    draw=violet!70!black,dashed},
  calib/.style={-{Latex[length=1.8mm]},line width=.7pt,
    draw=green!45!black,densely dotted}
]
\node[data] (survey) {CMS training split\\weighted contexts};
\node[data,below=of survey] (profile) {Synthetic traveler\\profile and traits};
\node[gate,below=of profile] (scenario) {Condition controls\\weather / fare / delay};

\node[planner,right=13mm of survey] (plan) {\textbf{1. Plan}\\fixed / mock / live LLM};
\node[gate,right=of plan] (check) {\textbf{2. Validate}\\typed checks and repair};
\node[module,right=of check] (choice) {\textbf{3. Decide}\\departure / mode / route};
\node[module,right=of choice] (network) {\textbf{4. Load}\\shared zone network};
\node[memory,below=9mm of choice] (memory) {\textbf{5. Update state}\\memory / habit / experience};

\node[output,right=13mm of network] (metrics) {Holdout evaluation\\JSD / Wasserstein / switching};
\node[gate,below=of metrics] (audit) {Trace log\\prompts / hashes / revisions};
\node[output,below=of audit] (replay) {Deterministic replay\\frozen response pairs};

\begin{scope}[shift={(survey.north west)},x=1mm,y=1mm]
  \draw[iconbg=green!50!black] (0,0) circle (2.35);
  \draw[iconline=green!45!black] (-1.15,.75) ellipse (1.15 and .4);
  \draw[iconline=green!45!black] (-1.15,.75)--(-1.15,-.9)
    (1.15,.75)--(1.15,-.9) (-1.15,-.05) arc[start angle=180,end angle=360,x radius=1.15,y radius=.4]
    (-1.15,-.9) arc[start angle=180,end angle=360,x radius=1.15,y radius=.4];
\end{scope}
\begin{scope}[shift={(profile.north west)},x=1mm,y=1mm]
  \draw[iconbg=green!50!black] (0,0) circle (2.35);
  \draw[iconline=green!45!black] (0,.85) circle (.55);
  \draw[iconline=green!45!black] (-1.15,-1.15) .. controls (-.9,.15) and (.9,.15) .. (1.15,-1.15);
\end{scope}
\begin{scope}[shift={(scenario.north west)},x=1mm,y=1mm]
  \draw[iconbg=orange!75!black] (0,0) circle (2.35);
  \draw[iconline=orange!70!black] (-1.2,.25) .. controls (-1.1,1.1) and (-.25,1.05) .. (-.1,.55)
    .. controls (.25,1.3) and (1.2,.9) .. (1.15,.15) -- (-1.2,.15);
  \draw[iconline=orange!70!black] (-.75,-.35)--(-1,-1.05) (0,-.35)--(-.25,-1.05) (.75,-.35)--(.5,-1.05);
\end{scope}
\begin{scope}[shift={(plan.north west)},x=1mm,y=1mm]
  \draw[iconbg=violet!65!black] (0,0) circle (2.35);
  \draw[iconline=violet!65!black] (-1.15,-1.25) rectangle (1.15,1.25);
  \draw[iconline=violet!65!black] (-.62,-1.25)--(-.62,1.25) (-.25,.65)--(.75,.65)
    (-.25,.05)--(.75,.05) (-.25,-.55)--(.55,-.55);
\end{scope}
\begin{scope}[shift={(check.north west)},x=1mm,y=1mm]
  \draw[iconbg=orange!75!black] (0,0) circle (2.35);
  \draw[iconline=orange!70!black] (-1.05,-1.2) rectangle (1.05,1.05);
  \draw[iconline=orange!70!black] (-.45,1.05)--(-.45,1.4)--(.45,1.4)--(.45,1.05);
  \draw[iconline=orange!70!black,line width=.85pt] (-.65,-.15)--(-.15,-.65)--(.75,.45);
\end{scope}
\begin{scope}[shift={(choice.north west)},x=1mm,y=1mm]
  \draw[iconbg=blue!55!black] (0,0) circle (2.35);
  \draw[iconline=blue!55!black] (-1.35,-.55) rectangle (1.35,.35);
  \draw[iconline=blue!55!black] (-.75,.35)--(-.35,1)--(.65,1)--(1.05,.35);
  \draw[iconline=blue!55!black] (-.85,-.75) circle (.35) (.85,-.75) circle (.35);
\end{scope}
\begin{scope}[shift={(network.north west)},x=1mm,y=1mm]
  \draw[iconbg=blue!55!black] (0,0) circle (2.35);
  \draw[iconline=blue!55!black] (-1,-.75)--(0,.95)--(1,-.55)--(-1,-.75) (0,.95)--(.15,-.2);
  \fill[blue!55!black] (-1,-.75) circle (.28) (0,.95) circle (.28) (1,-.55) circle (.28) (.15,-.2) circle (.28);
\end{scope}
\begin{scope}[shift={(memory.north west)},x=1mm,y=1mm]
  \draw[iconbg=yellow!55!black] (0,0) circle (2.35);
  \draw[iconline=yellow!45!black,-{Latex[length=1mm]}] (1.15,.55) arc[start angle=25,end angle=205,radius=1.25];
  \draw[iconline=yellow!45!black,-{Latex[length=1mm]}] (-1.15,-.55) arc[start angle=205,end angle=385,radius=1.25];
\end{scope}
\begin{scope}[shift={(metrics.north west)},x=1mm,y=1mm]
  \draw[iconbg=cyan!55!black] (0,0) circle (2.35);
  \draw[iconline=cyan!55!black] (-1.25,-1)--(-1.25,1.1) (-1.25,-1)--(1.3,-1);
  \fill[cyan!45!black] (-.85,-.9) rectangle (-.35,-.1) (-.15,-.9) rectangle (.35,.5) (.55,-.9) rectangle (1.05,.95);
\end{scope}
\begin{scope}[shift={(audit.north west)},x=1mm,y=1mm]
  \draw[iconbg=orange!75!black] (0,0) circle (2.35);
  \draw[iconline=orange!70!black] (-.35,.35) circle (1.05);
  \draw[iconline=orange!70!black,line width=.9pt] (.4,-.4)--(1.25,-1.25);
\end{scope}
\begin{scope}[shift={(replay.north west)},x=1mm,y=1mm]
  \draw[iconbg=cyan!55!black] (0,0) circle (2.35);
  \draw[iconline=cyan!55!black,fill=cyan!18] (-.65,-1)--(-.65,1)--(1,.0)--cycle;
\end{scope}

\draw[arr] (survey)--(plan);
\draw[arr] (profile.east)-|([xshift=-3mm]plan.west)--(plan.west);
\draw[arr] (scenario.east)-|([xshift=-5mm]choice.south west)--(choice.south west);
\draw[arr] (plan)--(check);
\draw[arr] (check)--(choice);
\draw[arr] (choice)--(network);
\draw[arr] (network)--(metrics);
\draw[arr] (network.south)|-(memory.east);
\draw[feedback] (memory.west)-|node[pos=.25,below,font=\scriptsize]{next day}(plan.south);

\draw[calib] (survey.east) .. controls +(16mm,-18mm) and +(-18mm,-13mm) .. (choice.south);
\draw[arr,draw=orange!75!black] (check.south)|-(audit.west);
\draw[arr,draw=orange!75!black] (audit)--(replay);
\draw[arr,draw=cyan!55!black] (metrics)--(audit);

\node[group,fit=(survey)(profile)(scenario),label={[font=\scriptsize\bfseries]above:Inputs and controls}] (ginput) {};
\node[group,fit=(plan)(check)(choice)(network)(memory),label={[font=\scriptsize\bfseries]above:Closed traveler-day simulation}] (gloop) {};
\node[group,fit=(metrics)(audit)(replay),label={[font=\scriptsize\bfseries]above:Evaluation and replay}] (gout) {};
\end{tikzpicture}
}
\caption{\calm\ architecture. Green nodes provide training-derived and synthetic inputs; the center panel executes a closed traveler-day loop with optional LLM planning; orange nodes and paths mark feasibility, experimental controls, and traceability components. Solid arrows are runtime flows, the dashed arrow is cross-day feedback, and the dotted arrow denotes training-only ASC calibration. Holdout metrics provide empirical targets, while network and stress outputs support controlled sensitivity analysis.}
\label{fig:architecture}
\end{figure*}

\section{Framework}
\subsection{Traveler-day loop}
Each synthetic traveler has attributes $z_i$ and class-conditioned traits $\theta_i$ (time, cost, reliability, comfort, environment, habit, risk). Training survey rows are weighted-sampled to assign purpose, OD, departure anchor, and habitual mode; demographics are synthetic. For the matched experiments, purpose and OD are held fixed across arms so that observed differences are attributable to the planner, memory, and choice modules rather than changing demand contexts.

Figure~\ref{fig:architecture} shows the architecture. A planner (fixed choice set, deterministic mock, or live LLM) proposes activities and consideration sets. A typed checker enforces mandatory context-matched purpose/destination, positive durations, consistent windows, and nonempty candidates (at most two regenerations before deterministic repair). Stochastic departure/mode/route choice and a shared zone network close the day; memory and optional habit adaptation feed the next day.

For mode $m$ in consideration set $\mathcal{M}_{it}$,
\begin{equation}
P(m\mid i,t)=\frac{\exp(U_{imt})}{\sum_{j\in\mathcal{M}_{it}}\exp(U_{ijt})},
\end{equation}
where $U_{imt}$ combines an ASC with normalized travel time and cost, reliability, transfers, walking, comfort, environment, and habit. The seven fitted ASCs relative to subway are $(-0.572,-2.579,0.535,0.583,-2.709,-2.924)$ for bus, taxi, car, walk, bike, and commuter rail, all inside predeclared $[-5,5]$ bounds.

The zone network supplies a shared state across travelers and experimental arms. Road time follows a Bureau of Public Roads form, transit adds capacity-based crowding, and walk/bike use fixed multipliers. This design closes the traveler-day feedback loop, while the current 13-zone instantiation provides a controlled shared-state environment for evaluating network-mediated behavioral responses.

\subsection{Evaluation components}
Table~\ref{tab:evidence} summarizes the task-aligned evaluation design. The CMS holdout measures empirical aggregate fit; matched live-LLM arms isolate planner and memory effects; mock/network runs exercise controlled interventions; and supervised MNL provides an individual trip-prediction reference. Together, these components connect system-level generation, controlled comparison, and predictive evaluation through metrics appropriate to each task.

\begin{table}[t]
\caption{Evaluation design and objectives.}
\label{tab:evidence}
\centering\footnotesize
\begin{tabular}{p{0.21\columnwidth}p{0.25\columnwidth}p{0.34\columnwidth}}
\toprule
Evidence & Role & Focus\\
\midrule
Survey holdout & Aggregate target & Person-disjoint fit\\
Live LLM arms & Matched ablation & Planner/memory effects\\
Mock/network runs & Intervention tests & Controlled response\\
Supervised MNL & Trip prediction & Individual prediction\\
\bottomrule
\end{tabular}
\end{table}

\section{Data and Experimental Design}
\subsection{Data freeze and split}
We retrieve the official 2024 NYC CMS trip table (dataset \texttt{wgnh-qwsg}) \cite{nycdot2024cms}. After normalization, 110,691 supported seven-mode trips from 3,368 respondents remain (subway, bus, taxi, car, walk, bicycle, commuter rail); 5,229 ``other'' records (1.97\% weighted) are excluded rather than reassigned. A fixed respondent-ID hash assigns 2,382 people (78,487 trips) to training and 986 people (32,204 trips) to holdout with zero overlap (Table~\ref{tab:data}). All ASCs and context sampling use training only.

\begin{table}[t]
\caption{Frozen CMS data flow.}
\label{tab:data}
\centering\footnotesize
\begin{tabular}{lrr}
\toprule
Artifact & People & Trips\\
\midrule
Raw normalized table & 3,397 & 115,920\\
Supported seven-mode table & 3,368 & 110,691\\
Training split & 2,382 & 78,487\\
Holdout split & 986 & 32,204\\
Excluded ``other'' & --- & 5,229\\
\bottomrule
\end{tabular}
\end{table}

\subsection{Metrics and experiments}
The primary endpoint is natural-log Jensen--Shannon divergence \cite{lin1991divergence} between weighted holdout mode shares $q$ and simulated shares $p$:
\begin{equation}
\jsd(p,q)=\tfrac12 D_{\mathrm{KL}}(p\Vert r)+\tfrac12D_{\mathrm{KL}}(q\Vert r),\quad r=\tfrac12(p+q).
\end{equation}
Secondary metrics (purpose JSD, departure/trip-time $\wone$/IQR, OD total variation, feasibility, switching) are reported separately; no raw-scale composite is formed.

\textbf{E2 (calibration).} Ten common seeds, 100 travelers, four days; ASCs fitted by training-only simulated method-of-moments and then frozen. Arms: habit, uncalibrated/calibrated choice, uncalibrated/calibrated mock. Uncertainty uses a two-source clustered bootstrap over paired simulated trajectories and holdout respondents (conditional on fitted ASCs).

\textbf{E3 (live ablation).} Matched 75 travelers $\times$ four condition days (300 traveler-days per arm); arms: calibrated choice-only, live LLM without memory, LLM with memory, full hybrid. Inference uses \texttt{gpt-4o-mini}, temperature 0.2, and prompt \texttt{v2.0-cms-context-audit}. Each call is keyed by a prompt hash; offline replay reuses the frozen response and reruns downstream validation and simulation without an additional API call.

\textbf{E5 (stress).} Rain, transit delay, fare, and parking severity ladders for calibrated choice and mock; the prespecified expectation is nonincreasing exposed-mode share.

\textbf{E6 (supervised).} Survey-weighted $L_2$-regularized MNL on individual trips versus a training-share prior provides a complementary predictive reference.

\section{Results}
\subsection{Person-disjoint calibration transfers to holdout}
\begin{figure}[t]
\centering
\includegraphics[width=0.86\columnwidth]{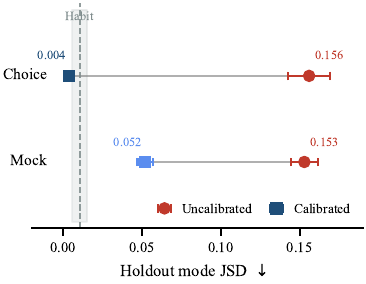}
\caption{E2 dumbbell plot: uncalibrated $\rightarrow$ calibrated holdout mode JSD (mean $\pm$ SD). Habit shown as a dashed reference.}
\label{fig:e2}
\end{figure}

Table~\ref{tab:e2} and Fig.~\ref{fig:e2} establish a stable empirical anchor for the framework. With all ASCs estimated on training respondents and then frozen, calibrated choice achieves mean holdout mode JSD 0.00394 versus 0.15599 before calibration (pooled difference $-0.15332$, 95\% interval $[-0.16577,-0.14057]$). The improvement persists across ten simulation seeds and transfers to respondents never used for fitting. Calibrated mock also improves substantially, while the habit arm provides an additional behaviorally simple reference at 0.01066.

\begin{table}[t]
\caption{E2 holdout mode JSD over ten seeds (lower is better).}
\label{tab:e2}
\centering\footnotesize
\begin{tabular}{lcc}
\toprule
Arm & Mean & SD\\
\midrule
Calibrated choice & \textbf{0.00394} & 0.00235\\
Habit & 0.01066 & 0.00480\\
Calibrated mock & 0.05205 & 0.00486\\
Uncalibrated mock & 0.15298 & 0.00849\\
Uncalibrated choice & 0.15599 & 0.01350\\
\bottomrule
\end{tabular}
\end{table}

The result provides \calm\ with a reproducible holdout baseline before any generative component is evaluated. It also shows the value of support alignment and person-disjoint validation: planner variants can be compared against the same frozen empirical target rather than against an in-sample aggregate.

\begin{table*}[!t]
\caption{E3 matched live-LLM ablation (300 traveler-days per arm). $\wone$ values are divided by holdout IQR.}
\label{tab:e3}
\centering\footnotesize
\begin{tabular}{lccccccc}
\toprule
Arm & Mode JSD$\downarrow$ & Depart.\ $\wone\downarrow$ & Trip-time $\wone\downarrow$ & Switch rate & First-pass feas.$\uparrow$ & Final feas.$\uparrow$ & Logged calls\\
\midrule
Choice-only & \textbf{0.00111} & \textbf{0.06173} & 1.33106 & 0.69778 & 1.00000 & 1.00000 & 0\\
LLM, no memory & 0.04515 & 0.07908 & 1.23185 & 0.47111 & 0.98667 & 1.00000 & 304\\
LLM + memory & 0.03595 & 0.07974 & 1.24148 & 0.36889 & 0.99333 & 1.00000 & 302\\
Full hybrid & 0.03838 & 0.08040 & \textbf{1.23121} & 0.32444 & 0.99667 & 1.00000 & 301\\
\bottomrule
\end{tabular}
\end{table*}

\subsection{Matched live-LLM ablation and deterministic replay}
\begin{figure}[t]
\centering
\includegraphics[width=0.76\columnwidth]{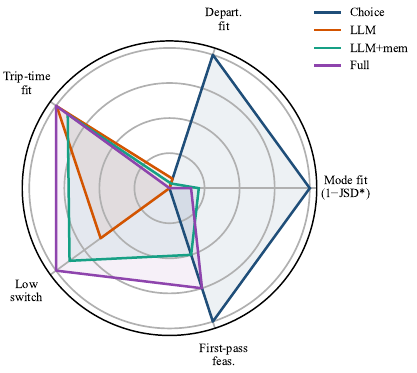}
\caption{E3 multi-metric profiles (radar; axes min--max normalized within the four arms so outer is relatively better).}
\label{fig:e3trade}
\end{figure}

Table~\ref{tab:e3} and Fig.~\ref{fig:e3trade} report the fully paired live experiment. Across the three live-LLM arms, all 900 traveler-days are rerun from frozen prompt--response pairs (614 unique pairs after cache reuse). First-pass feasibility exceeds 0.98 in every live arm and final feasibility is 1.0, demonstrating that typed validation and deterministic repair reliably integrate open-ended planner output into the simulator. Prompt hashes and frozen responses enable deterministic replay of the downstream simulation without additional model calls.

The matched design reveals a clear multi-objective trade-off. Choice-only retains the lowest mode JSD, whereas all live-LLM configurations reduce within-traveler switching and improve trip-time $\wone$/IQR; memory further lowers switching from 0.47111 to 0.36889, and the full hybrid reaches 0.32444. Because purpose and OD are fixed across arms, these differences isolate the behavior of the planner and memory modules. Thus, \calm\ does more than host a generative planner: it quantifies where that planner changes behavioral persistence and temporal fit while preserving feasibility and replay.

\subsection{Controlled intervention response}
\begin{figure}[t]
\centering
\includegraphics[width=\columnwidth]{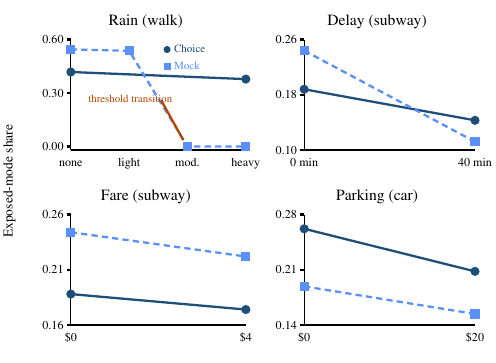}
\caption{E5 intervention responses. Choice curves connect the reported baseline and maximum-severity endpoints; the mock-rain path additionally shows the observed light$\rightarrow$moderate threshold transition.}
\label{fig:stress}
\end{figure}

Figure~\ref{fig:stress} and Table~\ref{tab:stress} show that calibrated choice responds directionally under rain, delay, fare, and parking ladders. All mean curves follow the prespecified monotone direction, confirming that intervention parameters propagate through choice and shared network state. The mock planner additionally produces a sharp walking transition under moderate rain (0.537 to 0.000). Capturing this transition alongside the expected directional response demonstrates that the stress protocol can characterize threshold-sensitive planner behavior beyond aggregate monotonic trends.

\begin{table}[t]
\caption{E5 mean exposed-mode share, baseline $\rightarrow$ maximum severity.}
\label{tab:stress}
\centering\footnotesize
\begin{tabular}{lcc}
\toprule
Scenario (exposed mode) & Choice & Mock\\
\midrule
Rain (walk) & 0.418 $\rightarrow$ 0.378 & 0.545 $\rightarrow$ 0.000\\
Delay (subway) & 0.188 $\rightarrow$ 0.143 & 0.244 $\rightarrow$ 0.112\\
Fare (subway) & 0.188 $\rightarrow$ 0.174 & 0.244 $\rightarrow$ 0.222\\
Parking (car) & 0.262 $\rightarrow$ 0.208 & 0.189 $\rightarrow$ 0.154\\
\bottomrule
\end{tabular}
\end{table}

\noindent\begin{minipage}{\columnwidth}
\subsection{Complementary supervised comparison}
On individual trip-mode prediction, weighted MNL obtains 0.5928 accuracy (macro-F1 0.2742; log loss 1.0337; Brier 0.5383) versus a training-share prior (0.3970 / 0.0880 / 1.4300 / 0.7170). The result supplies a competent predictive reference on the individual-choice task, complementing the simulator's aggregate-distribution evaluation.
\end{minipage}

\section{Discussion}
\calm\ connects three capabilities that are usually evaluated separately: empirical calibration, generative traveler planning, and shared-state simulation. Relative to traditional ABS, it introduces an optional generative planner behind typed feasibility and a frozen holdout target. Relative to mode-only or diary LLM systems \cite{wang2024urban,ju2025trajllm,liu2024limited}, it adds matched controls, disruption knobs, and replayable traces. Relative to persona-aligned choice models \cite{liu2026aligning}, it places choice within a multi-day loop in which memory and network state can affect subsequent decisions.

The experiments demonstrate why this integration matters. Person-disjoint calibration establishes the empirical anchor; matched live ablation reveals distinct trade-offs across aggregate fit, temporal fit, and behavioral persistence; and stress ladders verify directional response while characterizing threshold-sensitive behavior. These are complementary views of the same modular system. The result is an evaluation workflow in which generative components can be compared, reproduced, and improved without changing the surrounding simulator or empirical target.

\section{Limitations and Future Work}
The current study focuses on matched purpose--OD contexts and a controlled 13-zone network, enabling precise module-level comparisons. Its empirical conclusions therefore center on distributional calibration, matched planner effects, and directional intervention response within this setting. Extending the evaluation to endogenous activity-chain generation, empirically calibrated network conditions, observed disruption episodes, cross-fitted temporal and geographic holdouts, and additional cities represents the next stage of validation. These extensions build directly on the central contribution of the current study: a calibrated and reproducible scaffold in which each new generative or network component can be evaluated under matched conditions.

\section{Conclusion}
\calm\ provides a calibrated, reproducible, and modular framework for integrating LLM planning into activity-based traveler simulation. On NYC CMS, frozen training-only calibration transfers to a person-disjoint holdout, reducing mean mode JSD from 0.15599 to 0.00394 across ten seeds. Matched live experiments achieve final feasibility of 1.0, replay all 900 live-LLM traveler-days from frozen responses, and quantify trade-offs among mode fit, trip-time fit, and behavioral persistence. Controlled stress ladders produce interpretable directional responses and demonstrate the framework's ability to characterize planner sensitivity under changing travel conditions. Together, these results establish \calm\ as both a traveler-simulation architecture and a task-aligned evaluation protocol for generative mobility modules.

\IEEEtriggeratref{10}
\bibliographystyle{IEEEtran}
\bibliography{references}

\end{document}